\documentclass[final]{cvpr}

\usepackage{times}
\usepackage{epsfig}
\usepackage{graphicx}
\usepackage{amsmath}
\usepackage{amssymb}
\usepackage{multirow}

\usepackage[pagebackref=true,breaklinks=true,colorlinks,bookmarks=false]{hyperref}

\def\cvprPaperID{2336} 
\def\confYear{CVPR 2021}
\usepackage{bbm}
\begin{document}

\title{Can Semantic Labels Assist Self-Supervised Visual Representation Learning?}

\author{
Longhui Wei\textsuperscript{1,2}\quad Lingxi Xie\textsuperscript{2}\quad Jianzhong He\textsuperscript{2}\quad Jianlong Chang\textsuperscript{2}\\
Xiaopeng Zhang\textsuperscript{2}\quad Wengang Zhou\textsuperscript{1}\quad Houqiang Li\textsuperscript{1}\quad Qi Tian\textsuperscript{2}\\
\textsuperscript{1}University of Science and Technology of China,\quad\textsuperscript{2}Huawei Inc.\\
\small\texttt{weilh2568@gmail.com}\quad\small\texttt{198808xc@gmail.com}\quad\small\texttt{\{jianzhong.he,jianlong.chang\}@huawei.com}\\
\small\texttt{zxphistory@gmail.com}\quad\small\texttt{\{zhwg,lihq\}@ustc.edu.cn}\quad\small\texttt{tian.qi1@huawei.com}
}

\maketitle

\begin{abstract}
Recently, contrastive learning has largely advanced the progress of unsupervised visual representation learning. Pre-trained on ImageNet, some self-supervised algorithms reported higher transfer learning performance compared to fully-supervised methods, seeming to deliver the message that human labels hardly contribute to learning transferrable visual features. In this paper, we defend the usefulness of semantic labels but point out that fully-supervised and self-supervised methods are pursuing different kinds of features. To alleviate this issue, we present a new algorithm named \textbf{Supervised Contrastive Adjustment in Neighborhood} (SCAN) that maximally prevents the semantic guidance from damaging the appearance feature embedding. In a series of downstream tasks, SCAN achieves superior performance compared to previous fully-supervised and self-supervised methods, and sometimes the gain is significant. More importantly, our study reveals that semantic labels are useful in assisting self-supervised methods, opening a new direction for the community.
\end{abstract}

\section{Introduction}

Self-supervised learning has shown the potential of in various fields of artificial intelligence. In the natural language processing (NLP) community, researchers have shown a promising pipeline that uses a large amount of unlabeled data for model pre-training (\textit{e.g.}, BERT~\cite{devlin2018bert} and GPT-3~\cite{brown2020language}) and then fine-tunes them in a series of downstream tasks. This idea is attracting increasing attentions in the computer vision community, yet the key is to define proper pretext tasks (\textit{e.g.}, solving jigsaw puzzles~\cite{noroozi2016unsupervised,wei2019iterative,doersch2015unsupervised}, colorization~\cite{zhang2016colorful}, \textit{etc.}) to facilitate learning efficient visual representations from unlabeled image data.


Recently, this progress is largely advanced by contrastive learning~\cite{he2020momentum,chen2020simple,chen2020improved,tian2019contrastive,tian2020makes,grill2020bootstrap}. The assumption is that each image can be encoded into a low-dimensional feature vector and the encoding scheme is resistant against data augmentation (\textit{e.g.}, image cropping and horizontal flipping) so that the algorithm can distinguish the instance from a large gallery containing its variant as well as other instances. Contrastive learning has claimed significant improvement over traditional pretext tasks and, interestingly, the performance in some popular downstream tasks (including object detection and image segmentation, see Tab.~\ref{tab:0}) has surpassed the fully-supervised counterpart. This raises an important question: are the semantic labels useless for transferring the learned visual representations to other scenarios?

\begin{table}[!t]
    \caption{Comparison among three pre-training methods, namely, fully-supervised (FSup), MoCo (v2~\cite{chen2020improved}), and SCAN (our approach), in the performance (\%) of downstream transfer. ResNet-50 is used as the standard backbone, and all models are pre-trained only on ImageNet-1K. SCAN enjoys advantages over a wide range of datasets, tasks, and evaluation protocols. Please refer to the experimental section for full results. }
    \centering
    \setlength{\tabcolsep}{0.16cm}
    \begin{tabular}{|l|l|l||c|c|c|}
        \hline
        \multirow{2}*{Dataset} & \multirow{2}*{Task} & \multirow{2}*{Metric} & \multicolumn{3}{c|}{Pre-Training Method}  \\
        \cline{4-6}
         & & & FSup & MoCo & \textbf{SCAN} \\
        \hline\hline
        \multirow{2}*{VOC} & Det & AP$_{50}$ & 81.4 & 82.5 & \textbf{83.3} \\
        \cline{2-6}
        & Seg & mIoU & 74.4 & 73.4 & \textbf{76.6} \\
        \hline
        \multirow{2}*{COCO} & Det & AP$^\mathrm{bb}$ & 38.9 & 39.2 & \textbf{40.9} \\
        \cline{2-6}
         & InsSeg & AP$^\mathrm{mk}$ & 35.4 & 35.7 & \textbf{37.2} \\
        \hline
        LVIS & InsSeg & AP & 24.4 & 25.3 & \textbf{25.4} \\
        \hline
        \multirow{2}*{Cityscapes} & InsSeg & AP & 32.9 & 33.1 & \textbf{33.8} \\
        \cline{2-6}
         & Seg & mIoU & 74.6 & 75.2 & \textbf{76.5} \\
        \hline
    \end{tabular}
    \label{tab:0}
\end{table}

\begin{figure}[!t]
    \centering
    \includegraphics[width=0.48\textwidth]{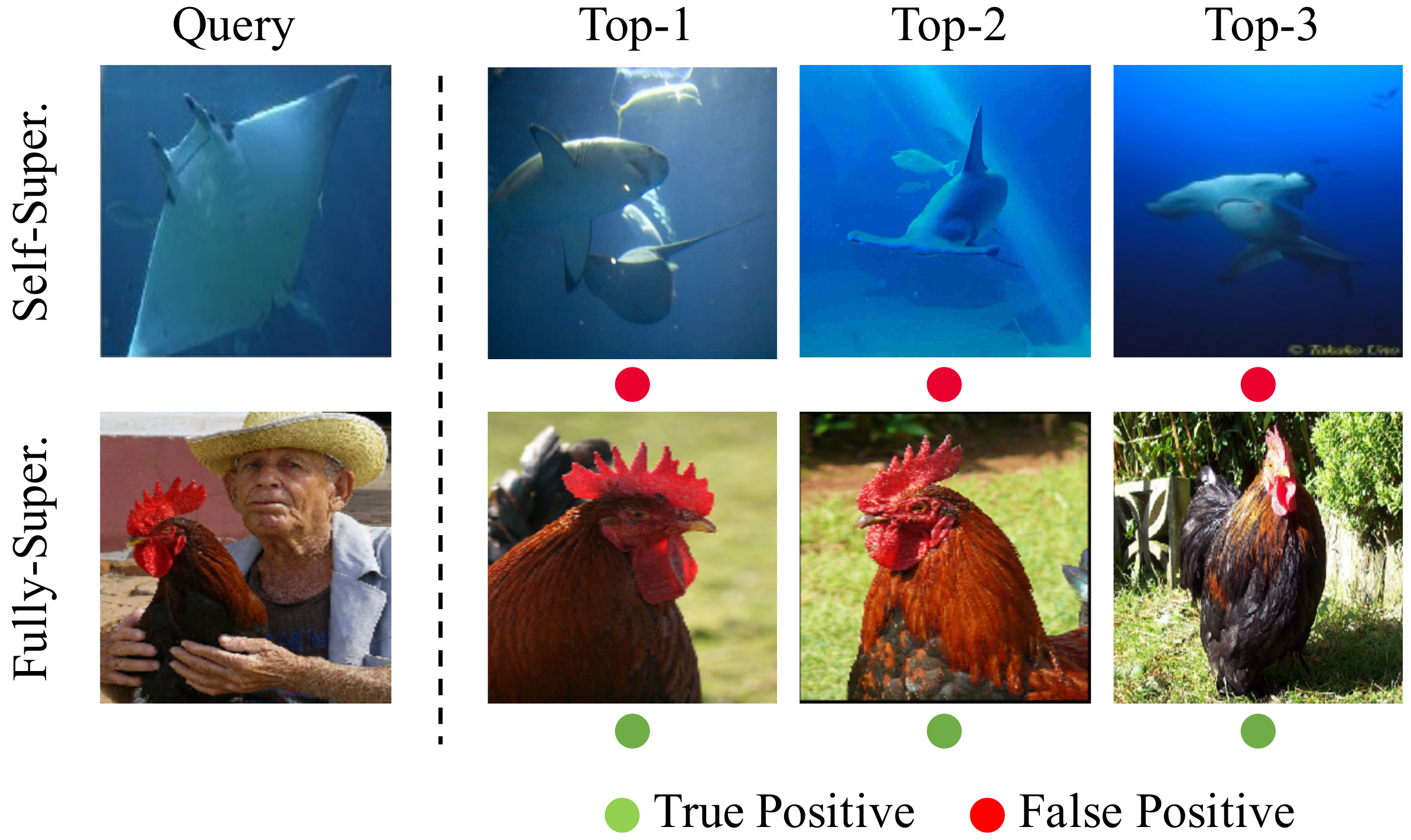}
    \caption{Examples of retrieval results with the features generated by self-supervised (MoCo-v2) and traditional fully-supervised method (trained with Softmax loss). In each example, the top 3 most similar images are displayed. True Positive means that the returned images belong to the same class of the query, and False Positive implies different classes. Clearly, self-supervised learning tends to extract global appearance information but ignore the details. Differently, fully-supervised learning focuses on the limited semantic regions.}
    \label{fig:fig1}
\end{figure}

This paper aims to answer the above question and defend the usefulness of semantic labels in visual representation learning. We first point out that fully-supervised and self-supervised methods are trying to capture different kinds of features for visual representation. We refer to them as \textbf{task-specific semantic features} and \textbf{task-agnostic appearance features}, respectively, where the former mainly focuses on semantic understanding and the latter is better at instance discrimination. Typical examples are shown in Figure~\ref{fig:fig1} where we extract features from a query image and try to find its nearest neighbors in the gallery. When the semantic features are used, the returned images are mostly from the same object class but the general layout may be quite different; when the appearance features are used, the opposite phenomenon is observed. Though the appearance features are verified to have better transfer ability, we expect the semantic features to assist visual understanding, \textit{e.g.}, improving the classification scores in detection and segmentation. This motivates us to design an algorithm that absorbs the advantages of both kinds of features.

However, we shall be aware that if fully-supervised and self-supervised objectives are directly combined, they may conflict with each other and thus deteriorate the performance of downstream tasks. To avoid the conflict, we present a novel algorithm named \textbf{Supervised Contrastive Adjustment in Neighborhood} (SCAN). The core part is a new definition of image-level relationship. Note that fully-supervised learning assumes that all images from the same class shall produce similar feature representations, while contrastive learning aims to distinguish all different images. SCAN is kind of compromise between them, trying to \textbf{enhance the relationship between each instance and others that are close to it in both semantics and appearance.}

In practice, SCAN is implemented by inheriting a pre-trained model by unsupervised contrastive learning, \textit{e.g.}, MoCo~\cite{chen2020improved}. Then, for each query, instead of treating all other images in the gallery as negative samples, SCAN preserves several positive samples that are similar to the query in both semantics and appearance. The contrastive learning objective is hence modified so that positive samples are pulled towards each other while the negative samples are pushed away as usual. The required efforts in code editing beyond the existing contrastive learning methods are minor.


We evaluate SCAN on a wide range of downstream tasks for detection and segmentation. As shown in Tab.~\ref{tab:0}, SCAN outperforms the fully-supervised and self-supervised baselines consistently, sometimes significantly. In particular, though the state-of-the-art self-supervised learning methods~\cite{chen2020simple,he2020momentum} claimed the benefits from extensive (around $1$ billion) unlabeled images, SCAN is able to surpass their performance by training on labeled ImageNet-1K with around $1$ million images. The contributions of this paper can be summarized as follows:

\begin{itemize}
    \item
    To the best of our knowledge, this is the first work to reveal the conflict between self-supervised and fully-supervised representation learning.
    \item
    We present an effective method, SCAN, to relieve the conflict. Without extra computational overheads, SCAN significantly improves the transfer performance in a series of downstream tasks.
    \item
    More importantly, we defend the usefulness of semantic labels in visual pre-training. This paves a new direction for the community to further enhance the ability of pre-training models. 
\end{itemize}

\section{Related Work}
\label{sec:2}
Our work is proposed under the current state-of-the-art self-supervised and fully-supervised learning methods. In this section, we mainly summarize the existing methods of the above aspects and elaborate the differences between these methods and our approach.

\subsection{Self-Supervised Learning Methods}
Self-supervised learning can be regarded as one kind of unsupervised learning task, and the key to solving it is how to design a pretext task. In the early stage, some novel handcrafted pretext tasks were proposed to extract useful knowledge. For example, some works~\cite{vincent2008extracting,pathak2016context,zhang2016colorful,zhang2017split,kingma2013auto,rezende2014stochastic} designed the image restoration related pretext tasks, and utilized auto-encoding based approaches to recover the input image. Other handcrafted pretext tasks,~\emph{e.g.}, jigsaw puzzle~\cite{noroozi2016unsupervised,wei2019iterative,doersch2015unsupervised}, image rotation prediction~\cite{gidaris2018unsupervised,chen2019self}, \textit{etc}, have also been proposed to further push forward the final performance.

Recently, discrimination related methods (more specifically, contrastive learning methods) have achieved great success in the self-supervised learning field~\cite{wu2018unsupervised,gutmann2010noise,zhuang2019local,misra2020self,oord2018representation,henaff2019data,tian2019contrastive,hjelm2018learning}. Non-parametric Instance Discrimination~\cite{wu2018unsupervised} utilized the novel instance-level discrimination idea to extract useful knowledge from extensive unlabeled images. Moreover, Contrastive Predictive Coding~\cite{oord2018representation} adopted a probabilistic contrastive loss to predict future samples in the latent space. Contrastive Multiview Coding~\cite{tian2019contrastive} successfully extended the two views to more views of contrastive learning. Momentum Contrast (MoCo)~\cite{he2020momentum} built a dynamic but representation-consistent memory bank for contrastive learning. SimCLR~\cite{chen2020simple} evaluated that larger batch size and suitable composition of data augmentation operations have clear benefits to the current contrastive learning framework. Moreover, the work~\cite{tian2020makes} empirically analyzed the importance of view selection.

The appealing points for self-supervised learning not only lie on they can extract useful knowledge from unlabeled data, but also they have evaluated that even trained on the same data, they still benefit more to downstream tasks compared with previous fully-supervised learning methods. Differently, our proposed method targets to reveal the opposite conclusion that, with rationally leveraging the human annotations, we can achieve much better performance of pre-training models on downstream tasks.

\subsection{Fully-Supervised Learning Methods}
Except for the network architecture design, fully-supervised learning methods generally contain two aspects: data augmentation methods and loss functions. Multiple novel data augmentation methods have been proposed in previous years,~\emph{e.g.}, Cutout~\cite{devries2017improved}, Mixup~\cite{zhang2017mixup}, CutMix~\cite{yun2019cutmix}, AutoAugment~\cite{cubuk2018autoaugment} and RandAugment~\cite{cubuk2020randaugment}. As for loss functions, Softmax loss is the most commonly used loss function. Moreover, label smoothing~\cite{muller2019does} and knowledge distillation related loss functions~\cite{hinton2015distilling,wei2020circumventing} are also widely used as the auxiliary loss. Recently, Supervised Contrastive Learning (SCL) has successfully extended the contrastive loss~\cite{khosla2020supervised} from self-supervised learning fields to fully-supervised learning fields, and it achieved competitive performance compared with the traditional Softmax loss. Though fully-supervised learning methods have achieved excellent performance on target tasks, the learned semantic features are biased and cannot transfer well into downstream tasks. Differently, our approach targets to learn a general pre-training feature space with the guidance of semantic labels and self-supervised learning methods.

\section{Our Method}
\subsection{Preliminaries: Momentum Contrast}
Momentum Contrast (MoCo) is a current state-of-the-art self-supervised learning method. Similar to most of the previous methods~\cite{tian2019contrastive,wu2018unsupervised,chen2020simple}, MoCo also chooses the instance discrimination mechanism as its pretext task. Differently, MoCo builds a memory bank for enlarging the look-up dictionary, and maintains an additional encoder by momentum update policy to dynamically update features in the memory bank. Through the above strategy, MoCo can nearly ignore the limitation of GPU memory, and compare a query sample with countless queue samples in the memory bank. Therefore, suppose a query image as $\mathbf{q}_i$, its two corresponding augmented images are denoted as $\mathbf{q}_i^{\dag}$ and $\mathbf{q}_i^{\ddag}$, the encoders for the query and memory bank are $\mathbf{f}$ and $\mathbf{g}$, the memory bank size is $L$, and the feature of each sample embedded in this memory bank is represented as $\mathbf{z}_l$, the loss function of MoCo for this query image can be formulated as:
\begin{equation}\label{eq:1}
    \mathcal{L}^\mathrm{MoCo}_i = -\log\frac{\exp{\{\mathbf{f}(\mathbf{q}_i^{\dag})^\top\cdot\mathbf{g}(\mathbf{q}_i^{\ddag})/\tau}\}}{\sum_{l=1}^L\exp{\{\mathbf{f}(\mathbf{q}_i^{\dag})^\top\cdot\mathbf{z}_l/\tau\}}},
\end{equation}
where $\tau$ represents the temperature hyper-parameter. More details about MoCo can be seen in this work~\cite{he2020momentum}.

As shown in Eq.~\eqref{eq:1}, the goal of MoCo is to pull in the same instance with different transformations, and enlarge the distances among the query and samples in the memory bank.
With this simple but effective scheme, MoCo heavily advances the development of self-supervised learning field.

\begin{figure*}[!t]
    \centering
    \includegraphics[width=0.98\textwidth]{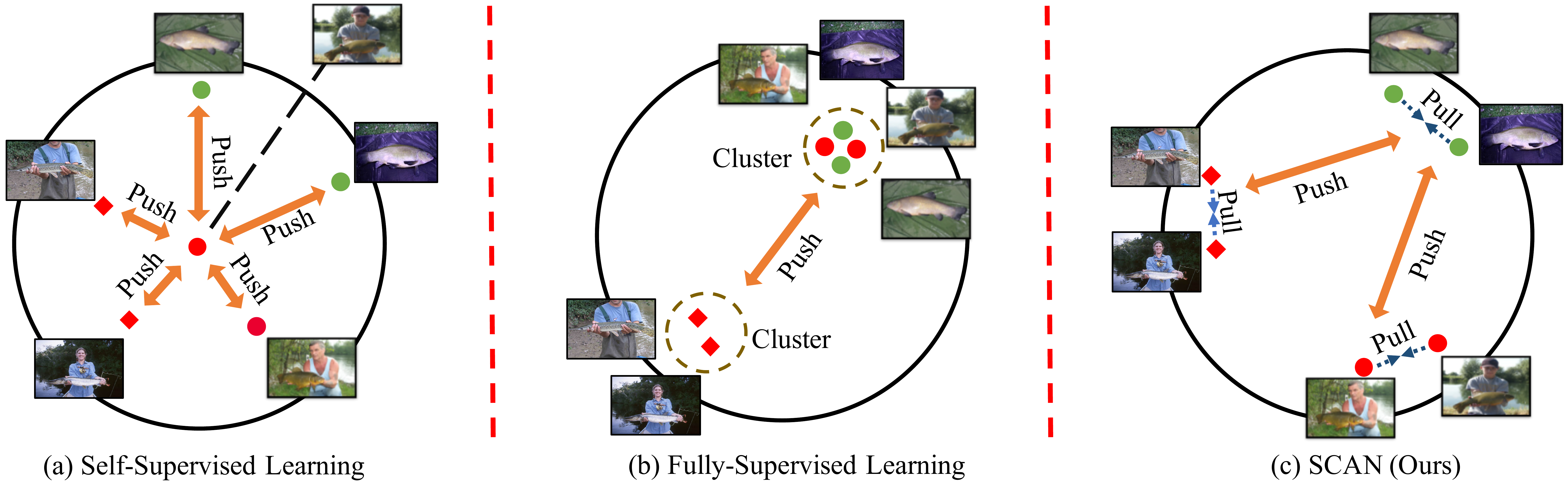}
    \caption{Illustrations of representation learning for self-supervised learning (instance discrimination based), fully-supervised learning, and our approach. Different icon shapes denote different classes, and different icon colors represent the corresponding images are dissimilar in global appearance. Different from self-supervised learning aiming to achieve instance-level discrimination and fully-supervised learning targeting to cluster the images of the same annotated class, SCAN tries to pull appearance-similar images of the same class closer and push other images away. }
    \label{fig:fig2}
\end{figure*}

\subsection{Conflict between Fully-Supervised and Self-Supervised Learning}
Compared with previous fully-supervised learning methods, MoCo achieves superior performance in downstream tasks,~\emph{e.g.}, $1.1\%$ improvement on AP$_{50}$ metric of detection task in VOC. Though MoCo has evaluated self-supervised learning methods can learn more transferrable knowledge, it is still doubtful that extra human annotations show no advantages for pre-training models. Compared with the raw images, human annotations can be regarded as extra semantic information. Therefore, the labeled dataset provides more knowledge than the unlabeled data, and in theory, the learned knowledge by self-supervised learning methods should be one subset of that learned by fully-supervised learning methods. However, the existing experiments reveal an opposite conclusion that using semantic labels harms the downstream performance. To investigate this paradox, we delve deep into the representation learning natures in the self-supervised and fully-supervised learning setting.

Interestingly, we find self-supervised and fully-supervised learning are pursuing different kinds of features, \textbf{task-agnostic appearance feature} and \textbf{task-specific semantic feature}. As shown in Fig.~\ref{fig:fig2}, for each query image, these self-supervised learning methods based on contrastive learning (like MoCo) target to push away all different samples for achieving the goal of instance-level discrimination. Consequently, they tend to extract the global appearance feature for roughly describing the whole of image. In their feature spaces, the appearance-similar images will locate closer and the appearance-dissimilar images will be far away. Therefore, the learned knowledge by self-supervised approaches is nearly task-agnostic, and can be well transferred into downstream tasks. However, as shown in Fig.~\ref{fig:fig1}, the generated appearance feature hardly to describe the semantic details, which reflects there is still much room to improve. Differently, fully-supervised learning methods pull images of the same class into one cluster and push images of the different classes away, ignoring other cues. Hence, fully-supervised algorithms will always attend on the limited task-relevant regions. Though the final generated semantic feature is more discriminative, it is heavily biased and cannot ensure the performance on every downstream task.


To summarize, the representation learning goal of self-supervised and fully-supervised learning methods are totally different, and there are both advantages and weaknesses of them for visual pre-training task.
Therefore, how to relieve the conflict of representation learning on them and effectively exploit their unique superiority is a good direction to improve the pre-training models. We shall present our solution in the next part.

\subsection{Our Solution: Supervised Contrastive Adjustment in Neighborhood}\label{scan}

The aforementioned conflict mainly comes from the inconsistency between the semantic and appearance spaces -- some samples are similar in semantics but not in appearance, and the opposite situation can also happen. To maximally avoid it, we design a conservative method that only takes the intersection of the semantic and appearance neighborhoods into consideration. Our algorithm is named \textbf{Supervised Contrastive Adjustment in Neighborhood} (SCAN). As shown in Fig.~\ref{fig:fig2}, SCAN is surprisingly simple: an extra judgment is added beyond the semantic labels, filtering the images sharing the same class label but the appearance is not sufficiently close to the query. Then, these remained images similar in both semantics and appearance to the query are pulled together, and other images are pushed away. In this way, SCAN maximally keeps the consistency between semantics and appearance similarity, and learns both kinds of knowledge simultaneously.

Therefore, the most important task for SCAN is to find the images that are similar to each other in both semantics and appearance.
Mathematically, for each image pair, $\mathbf{x}_i$ and $\mathbf{x}_{j}$, the semantic similarity of them can be simply measured by the corresponding semantic labels, as follows:
\begin{equation}\label{eq:-1}
\mathrm{Sim_S}(\mathbf{x}_i,\mathbf{x}_j)= \begin{cases} 1,& \quad\mathrm{if}\ \mathbf{y}_i=\mathbf{y}_j,\\
0,& \quad\mathrm{otherwise}, \end{cases}
\end{equation}
where $\mathbf{y}_i$ denotes the class label of $\mathbf{x}_i$. Since there is no ground-truth for measuring the appearance similarity, we refer to any pre-trained models that have not used semantic labels (\textit{e.g.}, MoCo-v2) for feature extraction. Let $\mathbf{a}_i$ and $\mathbf{a}_j$ be the features of $\mathbf{x}_{i}$ and $\mathbf{x}_{j}$, the appearance similarity of every two images is computed by:
\begin{equation}\label{eq:3}
    \mathrm{Sim_A}(\mathbf{x}_i,\mathbf{x}_j)= \frac{1}{2}\left(\mathbf{a}_i^\top\cdot\mathbf{a}_j+1\right).
\end{equation}
Note that the values of both $\mathrm{Sim_S}(\cdot)$ and $\mathrm{Sim_A}(\cdot)$ falls in the range of $[0,1]$, with larger values indicating higher similarity. Integrating Eq.~\eqref{eq:-1} and Eq.~\eqref{eq:3} yields the similarity measurement that takes both semantics and appearance into consideration:
\begin{equation}\label{eq:-2}
    \mathrm{Sim_{S\&A}}(\mathbf{x}_i,\mathbf{x}_j) = \mathrm{Sim_S}(\mathbf{x}_i,\mathbf{x}_j) \times \mathrm{Sim_A}(\mathbf{x}_i,\mathbf{x}_j).
\end{equation}
That being said, two images are considered similar only if they share similar layouts and the same semantic label.

Eq.~\eqref{eq:-2} offers a new way to generate the pseudo labels for contrastive learning. Given the query image, we compute its similarity to each candidate in the gallery using Eq.~\eqref{eq:-2}, and retrieve the top-ranked images with non-zero similarity as the positive samples (or called positive neighbors). This avoids the burden that two images with either distinct semantic or appearance features to be assigned the identical pseudo label, and thus alleviates the risk of conflict in the feature space.

Naturally, the generated pseudo labels embed well both the appearance and semantic information. Through simply pushing away images of the different assigned pseudo classes, SCAN inherits the ability of capturing appearance features. By pulling images of the same pseudo class together, SCAN is forced to embed semantic information into the feature space. In our implementation, limited by the GPU memory, SCAN also resorts to maintain a memory bank as MoCo for storing a large scale of samples appearing in the previous mini-batch, which will be regarded as negative samples in the current mini-batch. Therefore, suppose a total of $S$ images and their corresponding top-$K$ positive neighbors are sampled in one mini-batch, the corresponding loss function of SCAN for each query image $\mathbf{x}_i$ can be formulated as:
\begin{align}\label{eq:4}
    \mathcal{L}^\mathrm{SCAN}_i =  -\frac{1}{K} & \sum_{j=1}^{S(K+1)}\mathbbm{1}_{y_i^*=y_j^*} \nonumber \\ & -\log\frac{\exp{\{\mathbf{f}(\mathbf{x}_i^{\dag})^\top\cdot\mathbf{g}(\mathbf{x}_j^{\ddag})/\tau}\}}{\mathrm{Sim}^\mathrm{all}_i},
\end{align}
in which
\begin{align}\label{eq:6}
    \mathrm{Sim}^\mathrm{all}_i = \sum_{t=1}^{S(K+1)}\mathbbm{1}_{y_i^*\ne{y_t^*}} & \exp{\{\mathbf{f}(\mathbf{x}_i^{\dag})^\top\cdot\mathbf{g}(\mathbf{x}_t^{\ddag})/\tau\}} \nonumber \\ & +\sum_{l=1}^{L}\exp{\{\mathbf{f}(\mathbf{x}_i^{\dag})^\top\cdot\mathbf{z}_l/\tau\}}
\end{align}
where $\mathbbm{1}$ represents the indicator function, and $y_i^*$ denotes the generated pseudo label of $\mathbf{x}_i$.
It is worth noting that SCAN approximately degenerates to prior approaches, MoCo~\cite{he2020momentum} and supervised contrastive learning~\cite{khosla2020supervised} (SCL, injecting contrastive loss into fully-supervised learning), by setting the variable of $K$ in Eq.~\eqref{eq:4} to be $0$ and $\infty$, respectively. Though the formulations of these three methods are similar, the motivations of them are different. Unlike MoCo targeting to learn more useful knowledge from unlabeled data, or SCL aiming for achieving better performance on the image classification task, the core of SCAN is to improve the performance of visual pre-training models. Therefore, SCAN is designed to relieve the representation learning conflict between self-supervised learning and fully-supervised learning, and then maximally learns transferrable knowledge from both appearance and semantics. We refer the readers to Sec.~\ref{sec461} for further empirical analysis.


\section{Experiments}

\subsection{Datasets}
\label{sec:5.1}
We pre-train SCAN on ImageNet-1K~\cite{ImageNet}, the most widely used large-scale classification dataset. It contains about $1.28$M images of $1000$ classes. Because of the extensive annotations on ImageNet-1K, researchers usually train the backbones on this dataset and then fine-tune them on other downstream tasks for better performance. 

Following MoCo, we evaluate SCAN on four commonly used detection and segmentation datasets,~\emph{i.e.}, PASCAL VOC~\cite{everingham2010pascal}, COCO~\cite{lin2014microsoft}, LVIS~\cite{gupta2019lvis} and Cityscapes~\cite{cordts2016cityscapes}, respectively. PASCAL VOC and COCO involve both bounding boxes and segmentation annotations. Compared with PASCAL VOC, COCO is relatively larger and more complex. LVIS is a recently released long-tailed distributed instance segmentation dataset, which contains over $1000$ categories. In this paper, we use the recommended version, LVIS-v0.5, for evaluation. Additionally, Cityscapes is a widely used street scene segmentation dataset. In the following of this paper, we refer to PASCAL VOC as VOC for brevity.

\subsection{Implementation Details}
We directly utilize the official released MoCo-v2 model trained on ImageNet-1K with $800$ epochs to generate the appearance feature, and then compute the similarity of samples as Eq.~\eqref{eq:-2} to retrieve positive neighbors accordingly. Empirically, we assign each sample and its top-2 positive neighbors to the same new label. Finally, we select ResNet-50~\cite{he2016deep} as the used backbone and train it with the guidance of our generated labels. For one mini-batch of each GPU, we randomly choose $128$ samples and their corresponding positive neighbors. Therefore, there are always positive samples for each anchor image in the mini-batch. Similar to MoCo, we adopt the SGD optimizer and set momentum as $0.9$. Moreover, the cosine learning rate schedule is utiliezd, and the initial learning rate is set as $1.6$. The temperature $\tau$ in Eq.~\eqref{eq:4} is empirically set as $0.07$. Finally, we use $32$ Tesla-V100 GPUs to train our model lasting for $400$ epochs on ImageNet-1K.

\subsection{Results on PASCAL VOC}
\noindent\textbf{Experiments Setup.} For the detection task on VOC, we use Detectron2~\cite{wu2019detectron2} to fine-tune the Faster-RCNN~\cite{ren2015faster} with R50-C4 backbone. Same with MoCo, a multi-scale training policy is utilized and the testing image size is set as $800$. The overall of fine-tuning process lasts for $24\mathrm{K}$ iterations. All of the models are fine-tuned on VOC$_\mathrm{trainval07+12}$, and tested on VOC$_\mathrm{test2007}$. As for the semantic segmentation task, we use the same FCN-based structure~\cite{long2015fully} as MoCo to conduct the evaluations. All of the models are fine-tuned on VOC$_\mathrm{train_{aug}2012}$ and tested on VOC$_\mathrm{val2012}$. The training image size is set as $513$ and the original image size is used in the testing stage. To successfully reproduce the reported performance in MoCo, the training iterations are extended from $30\mathrm{K}$ to $50\mathrm{K}$. 

\begin{table}[]
    \centering
    \caption{Results of different pre-training methods on VOC. VOC$_\mathrm{Det}$/VOC$_\mathrm{Seg}$ means detection and semantic segmentation task on VOC, respectively. 
    IN denotes the pre-training dataset is ImageNet-1K, and IG represents the pre-training dataset is Instagram. FSup-Softmax means the backbone is trained with the Softmax loss under the fully-supervised setting.
    The reported numbers of SCAN are the average of 3 runs.}
    \scalebox{0.9}{
    \begin{tabular}{l|c|ccc|c}
        \hline
        \multirow{2}*{Methods} & \multirow{2}*{Data} & \multicolumn{3}{c|}{ VOC$_\mathrm{Det}$ } & \multicolumn{1}{c}{ VOC$_\mathrm{Seg}$ } \\
        \cline{3-6}
         & & AP$_{50}$ & AP$_{75}$ & AP & mIoU \\
        \hline\hline
        FSup-Softmax~\cite{he2020momentum} & IN & 81.4 & 58.8 & 53.5 & 74.4\\
        \hline
        PIRL~\cite{misra2020self} &IN & 81.1 & - & - & -\\
        MoCo-v1~\cite{he2020momentum} &IN & 81.5 & 62.6 & 55.9 & 72.5\\
        MoCo-v2~\cite{chen2020improved} &IN & 82.5 & 64.0 & 57.4 & 73.4\\
        MoCo-v1~\cite{he2020momentum} &IG & 82.2 & 63.7 & 57.2 & 73.6\\
        \hline
        SCAN &IN & \textbf{83.3} & \textbf{64.0} & \textbf{57.4} & \textbf{76.6}\\
        \hline
    \end{tabular}}
    \label{tab:1}
\end{table}

\noindent\textbf{Observations.}
The detection and semantic segmentation results on VOC are shown in Tab.~\ref{tab:1}. One interesting phenomenon is that, self-supervised and fully-supervised learning methods show different properties while transferring their learned knowledge into different downstream tasks. In the detection task, MoCo-v2 surpasses FSup-Softmax with a large margin,~\emph{e.g.}, $3.9\%$ improvement on Average Precision (AP) metric. However, there is $1.0\%$ performance degradation 
in the semantic segmentation task. Generally, the above results further reveal that, there exist both advantages and weaknesses of self-supervised and fully-supervised learning methods for the visual pre-training models.

Benefiting from effectively exploiting their unique advantages of self-supervised and fully-supervised learning methods, SCAN heavily pushes forward the final performances of pre-training models. Compared with the best of previously reported detection results of pre-training methods on VOC, there is $0.8\%$ improvement on AP$_{50}$ metric for SCAN. Moreover, compared with the best of previously reported semantic segmentation results on VOC, SCAN significantly improves it from \textbf{74.4\%} to \textbf{76.6\%}. The consistent improvements in different tasks sufficiently evaluate the effectiveness of our proposed method.

\begin{table*}[htb]
    \centering
    \caption{Results of utilizing different pre-training models to fine-tune Mask R-CNN with the R50-FPN backbone on COCO. COCO$_\mathrm{Det}$ and COCO$_\mathrm{InsSeg}$ denote the detection and instance segmentation task on COCO, respectively.}
    \scalebox{0.91}{
    \begin{tabular}{l|c|ccc|ccc||ccc|ccc}
        \hline
        \multirow{3}{*}{Methods}  & \multirow{3}{*}{Data}  &  \multicolumn{6}{c||}{Mask R-CNN,  R50-FPN, COCO$_\mathrm{Det}$} & \multicolumn{6}{c}{Mask R-CNN,  R50-FPN, COCO$_\mathrm{InsSeg}$} \\
        \cline{3-14}
        {} & {} &  \multicolumn{3}{c|}{1$\times$ schedule} & \multicolumn{3}{c||}{2$\times$ schedule} & \multicolumn{3}{c|}{1$\times$ schedule}  & \multicolumn{3}{c}{2$\times$ schedule} \\
        \cline{3-14}
        {} & {} &  \multicolumn{1}{c}{AP$^\mathrm{bb}$} & \multicolumn{1}{c}{AP$_{50}^\mathrm{bb}$} & \multicolumn{1}{c|}{AP$_{75}^\mathrm{bb}$} &
        \multicolumn{1}{c}{AP$^\mathrm{bb}$} & \multicolumn{1}{c}{AP$_{50}^\mathrm{bb}$} & \multicolumn{1}{c||}{AP$_{75}^\mathrm{bb}$} &
        \multicolumn{1}{c}{AP$^\mathrm{mk}$} & \multicolumn{1}{c}{AP$_{50}^\mathrm{mk}$} & \multicolumn{1}{c|}{AP$_{75}^\mathrm{mk}$} &
        \multicolumn{1}{c}{AP$^\mathrm{mk}$} & \multicolumn{1}{c}{AP$_{50}^\mathrm{mk}$} & \multicolumn{1}{c}{AP$_{75}^\mathrm{mk}$} \\
        \hline \hline
        FSup-Softmax~\cite{he2020momentum}& IN & 38.9 & 59.6& 42.7& 40.6& 61.3& 44.4& 35.4& 56.5& 38.1& 36.8& 58.1& 39.5\\
        \hline
        MoCo-v1~\cite{he2020momentum}& IN & 38.5 & 58.9& 42.0& 40.8& 61.6& 44.7& 35.1& 55.9& 37.7& 36.9& 58.4& 39.7\\
        MoCo-v2~\cite{chen2020improved}& IN & 39.2 & 59.9& 42.7& 41.6& 62.1& 45.6& 35.7& 56.8& 38.1& 37.7& 59.3& 40.6\\
        MoCo-v1~\cite{he2020momentum}& IG & 38.9 & 59.4& 42.3& 41.1& 61.8& 45.1& 35.4& 56.5& 37.9& 37.4& 59.1& 40.2\\
        \hline
        SCAN& IN & \textbf{40.9} & \textbf{61.9}& \textbf{44.7}& \textbf{42.3}& \textbf{63.1}& \textbf{46.3}& \textbf{37.2}& \textbf{58.8}& \textbf{39.9}& \textbf{38.2}& \textbf{60.2}& \textbf{41.1}\\
        \hline
    \end{tabular}}
    \label{tab:2}
\end{table*}

\begin{table*}[htb]
    \centering
    \caption{Results of utilizing different pre-training models to fine-tune Mask R-CNN with the R50-C4 backbone on COCO.}
    \scalebox{0.91}{
    \begin{tabular}{l|c|ccc|ccc||ccc|ccc}
        \hline
        \multirow{3}{*}{Methods}  &
        \multirow{3}{*}{Data}  &
        \multicolumn{6}{c||}{Mask R-CNN,  R50-C4, COCO$_\mathrm{Det}$} & \multicolumn{6}{c}{Mask R-CNN,  R50-C4, COCO$_\mathrm{InsSeg}$} \\
        \cline{3-14}
        {} & {} &  \multicolumn{3}{c|}{1$\times$ schedule} & \multicolumn{3}{c||}{2$\times$ schedule} & \multicolumn{3}{c|}{1$\times$ schedule}  & \multicolumn{3}{c}{2$\times$ schedule} \\
        \cline{3-14}
        {} & {} & \multicolumn{1}{c}{AP$^\mathrm{bb}$} & \multicolumn{1}{c}{AP$_{50}^\mathrm{bb}$} & \multicolumn{1}{c|}{AP$_{75}^\mathrm{bb}$} &
        \multicolumn{1}{c}{AP$^\mathrm{bb}$} & \multicolumn{1}{c}{AP$_{50}^\mathrm{bb}$} & \multicolumn{1}{c||}{AP$_{75}^\mathrm{bb}$} &
        \multicolumn{1}{c}{AP$^\mathrm{mk}$} & \multicolumn{1}{c}{AP$_{50}^\mathrm{mk}$} & \multicolumn{1}{c|}{AP$_{75}^\mathrm{mk}$} &
        \multicolumn{1}{c}{AP$^\mathrm{mk}$} & \multicolumn{1}{c}{AP$_{50}^\mathrm{mk}$} & \multicolumn{1}{c}{AP$_{75}^\mathrm{mk}$} \\
        \hline \hline
        FSup-Softmax~\cite{he2020momentum}& IN & 38.2 & 58.2& 41.2& 40.0& 59.9& 43.1& 33.3& 54.7& 35.2& 34.7& 56.5& 36.9\\
        \hline
        MoCo-v1~\cite{he2020momentum}& IN & 38.5 & 58.3& 41.6& 40.7& 60.5& 44.1& 33.6& 54.8& 35.6& 35.4& 57.3& 37.6\\
        MoCo-v2~\cite{chen2020improved}& IN & 39.5 & 59.1& 42.7& 41.2& 61.0& 44.8& 34.5& 55.8& 36.7& 35.8& 57.6& 38.3\\
        MoCo-v1~\cite{he2020momentum}& IG & 39.1 & 58.7& 42.2& 41.1& 60.7& 44.8& 34.1& 55.4& 36.4& 35.6& 57.4& 38.1\\
        \hline
        SCAN& IN & \textbf{40.1} & \textbf{60.2}& \textbf{43.2}& \textbf{41.7}& \textbf{61.7}& \textbf{45.4}& \textbf{34.9}& \textbf{56.6}& \textbf{37.1}& \textbf{36.2}& \textbf{58.3}& \textbf{38.6}\\
        \hline
    \end{tabular}}
    \label{tab:3}
\end{table*}

\subsection{Results on MS-COCO}\label{sec54}

\noindent\textbf{Experiments Setup.} COCO is the most popular large-scale detection and segmentation dataset, which includes multiple tasks,~\emph{e.g.}, object detection, instance segmentation, keypoint detection, \textit{etc}. In this section, we utilize Mask R-CNN~\cite{he2017mask} to conduct detection and instance segmentation task concurrently. Different backbones (R50-FPN and R50-C4) with different learning schedules (1$\times$ and 2$\times$ schedules) are used to evaluate the pre-training models. All of the above experiments are trained on COCO$_\mathrm{train2017}$ and tested on COCO$_\mathrm{val2017}$. 

\noindent\textbf{Observations.} The results of fine-tuning Mask R-CNN with different backbones on COCO are shown in Tab.~\ref{tab:2} and Tab.~\ref{tab:3}, respectively. Compared with FSup-Softmax, MoCo-v2 shows clear advantages in both detection and instance segmentation tasks, which seems to reveal that human annotations do not contribute to learning transferrable knowledge. However, with properly exploiting the available labels, SCAN can significantly outperform MoCo-v2,~\emph{e.g.}, $1.7\%$ improvement on AP metric for the R50-FPN backbone with $1\times$ schedule in COCO detection task, and $0.4\%$ improvement on AP metric for the R50-C4 backbone with $2\times$ schedule in COCO instance segmentation task, respectively. \textbf{The consistent improvements on COCO demonstrate the semantic labels do benefit pre-training models if they are exploited properly}.



\subsection{Results on Other Datasets}

\noindent\textbf{Experiments Setup.} We also conduct comparisons on some other special datasets,~\emph{e.g.}, LVIS-v0.5, and Cityscapes. Compared with COCO, LVIS-v0.5 is a more challenging instance segmentation dataset, which contains over 1000 long-tailed distributed categories. The long-tailed distribution problem makes the training process hard to converge. Different from VOC and COCO that consist of natural images, Cityscapes is a large-scale urban street scene segmentation dataset. Therefore, Cityscapes can better evaluate the robustness of different pre-training models trained on ImageNet-1K (consisting of natural images). For the instance segmentation task in LVIS-v0.5 and Cityscapes, we adopt Mask R-CNN with the R50-FPN backbone and 2$\times$ schedule to train on their corresponding 
training sets. As for the semantic segmentation task on Cityscapes, we use the same FCN-based model as MoCo to conduct the evaluations. All of the experimental settings are the same with MoCo, except that in the testing stage, we utilize slide inference policy as PSPNet~\cite{zhao2017pyramid} on Cityscapes for successfully reproducing the reported performance in MoCo.

\noindent\textbf{Observations.} The results on LVIS-v0.5 and Cityscapes are shown in Tab.~\ref{tab:4} and Tab.~\ref{tab:5}, respectively. Obviously, SCAN consistently outperforms MoCo-v2 on Cityscape,~\emph{e.g.}, $0.7\%$ improvement on AP metric in instance segmentation task and $1.3\%$ improvement on semantic segmentation task, respectively.
Though SCAN achieves nearly equal performance compared with MoCo-v2 on LVIS-v0.5, we can still conclude that SCAN is a safer scheme for training visual pre-training models.

\begin{table}[]
    \centering
    \caption{ Instance segmentation results of utilizing different pre-training models to fine-tune Mask R-CNN with the R50-FPN backbone and 2$\times$ schedule on LVIS-v05.}
    \scalebox{0.93}{
    \begin{tabular}{l|c|c|ccc}
        \hline
        \multirow{2}*{Methods} &
        \multirow{2}*{Data} &
        \multirow{2}*{BN} &\multicolumn{3}{c}{ LVIS-v05$_\mathrm{InsSeg}$ } \\
        \cline{4-6}
         & & & AP & AP$_{50}$ & AP$_{75}$ \\
        \hline\hline
        FSup-Softmax~\cite{he2020momentum} & IN & frozen & 24.4 & 37.8 & 25.8\\
        FSup-Softmax~\cite{he2020momentum} & IN & tuned & 23.2 & 36.0 & 24.4\\
        \hline
        MoCo-v1~\cite{he2020momentum}& IN & tuned & 24.1 & 37.4 & 25.5\\
        MoCo-v2~\cite{chen2020improved}& IN & tuned & 25.3 & 38.4 & \textbf{27.0}\\
        MoCo-v1~\cite{he2020momentum}& IG & tuned & 24.9 & 38.2 & 26.4\\
        \hline
        SCAN& IN & tuned & \textbf{25.4} & \textbf{38.8} & 26.8\\
        \hline
    \end{tabular}}
    \label{tab:4}
\end{table}

\begin{table}[]
    \centering
    \caption{ Results of different pre-training methods on instance and semantic segmentation task in Cityscapes. CS$_\mathrm{InsSeg}$ and CS$_\mathrm{Seg}$ represents the instance segmentation and semantic segmentation task in Cityscape, respectively. The reported results of SCAN are the average of 3 runs.}
    \scalebox{0.95}{
    \begin{tabular}{l|c|cc|c}
        \hline
        \multirow{2}*{Methods} &
        \multirow{2}*{Data} & \multicolumn{2}{c|}{ CS$_\mathrm{InsSeg}$ } & \multicolumn{1}{c}{ CS$_\mathrm{Seg}$ } \\
        \cline{3-5}
         & & AP & AP$_{50}$  &mIoU \\
        \hline\hline
        FSup-Softmax~\cite{he2020momentum}& IN  & 32.9 & 59.6 & 74.6\\
        \hline
        MoCo-v1~\cite{he2020momentum}& IN & 32.3 & 59.3 & 75.3\\
        MoCo-v2~\cite{chen2020improved}& IN & 33.1 & 60.1 & 75.2\\
        MoCo-v1~\cite{he2020momentum}& IG & 32.9 & 60.3 & 75.5\\
        \hline
        SCAN& IN  & \textbf{33.8} & \textbf{61.3} & \textbf{76.5}\\
        \hline
    \end{tabular}}
    \label{tab:5}
\end{table}

\subsection{Ablation Study}\label{sec56}
\subsubsection{The Impact of Hyper-parameters}\label{sec461}
As introduced in Sec.~\ref{scan}, the only hyper-parameter of our approach is the candidate top-$K$ positive neighbors. While setting the value of $K$ to be $0$, SCAN degenerates into a self-supervised learning formulation (MoCo). While seting the value of $K$ to be $\infty$, SCAN changes into a fully-supervised learning formulation (SCL). Therefore, selecting a suitable value of $K$ is important for SCAN.
\begin{figure}[!t]
    \centering
    \includegraphics[width=0.48\textwidth]{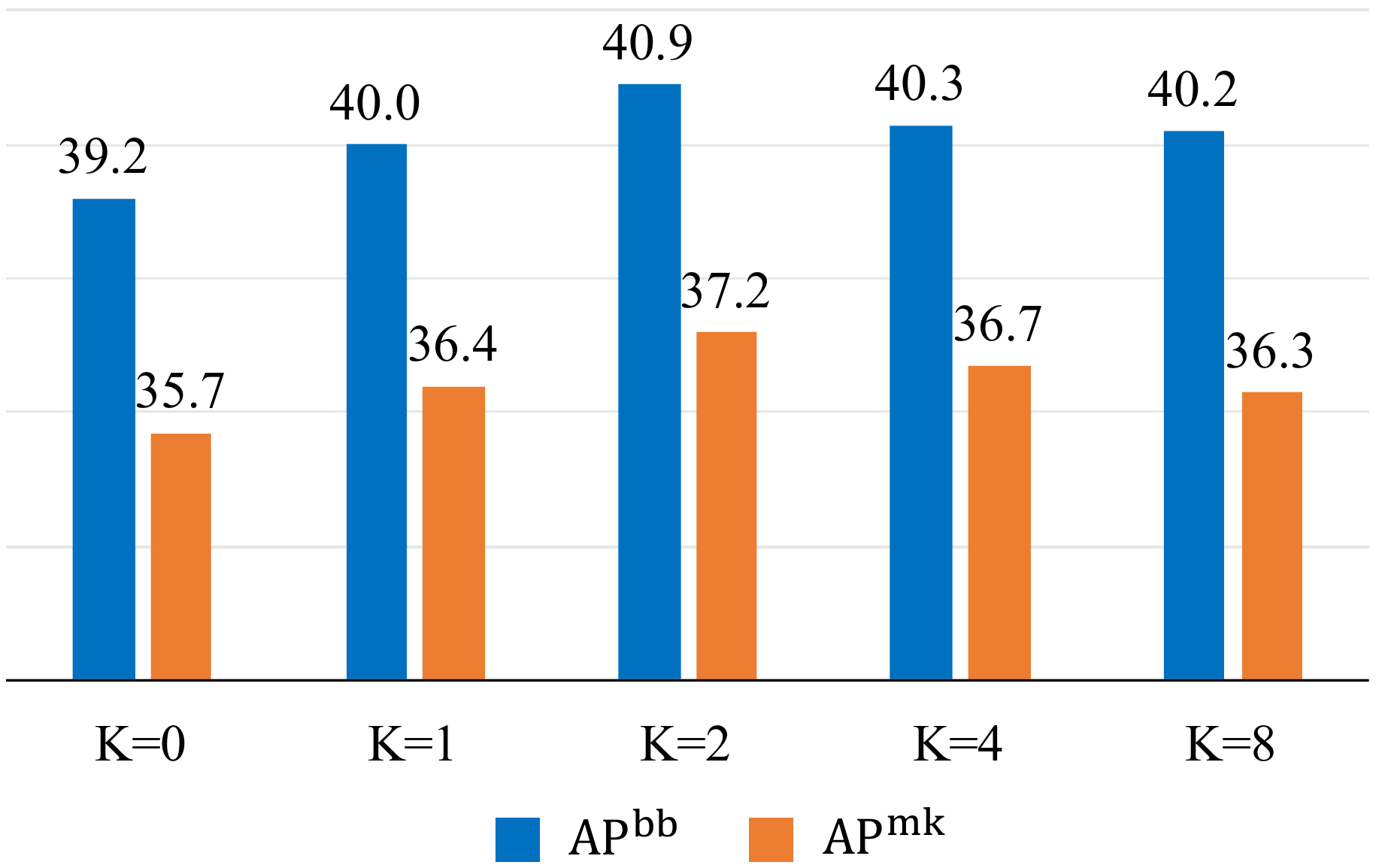}
    \caption{The impact of different values of $K$. $AP^{bb}$ and $AP^{mk}$ represents the Average Precision metric for detection and instance segmentation task on COCO by fine-tuning Mask R-CNN with the R50-FPN backbone and $1\times$ schedule, respectively.}
    \label{fig:fig4}
\end{figure}
As shown in Fig.~\ref{fig:fig4}, with enlarging $K$, the performances of SCAN on downstream tasks (detection and instance segmentation tasks on COCO) are firstly improved and then slowly declined. For example, with increasing the value of $K$ from $0$ to $2$, the detection performance of the pre-training model trained by SCAN is improved from $39.2\%$ to $40.9\%$ on COCO. That indicates enhancing the semantic discrimination ability is indeed required for pre-training models. However, while enlarging the value of $K$ from $2$ to $8$, the instance segmentation performance on COCO is decreased from $37.2\%$ to $36.3\%$, which reveals clustering appearance-dissimilar images of the same class will destroy the generalization ability.

\begin{figure}[!t]
    \centering
    \includegraphics[width=0.48\textwidth]{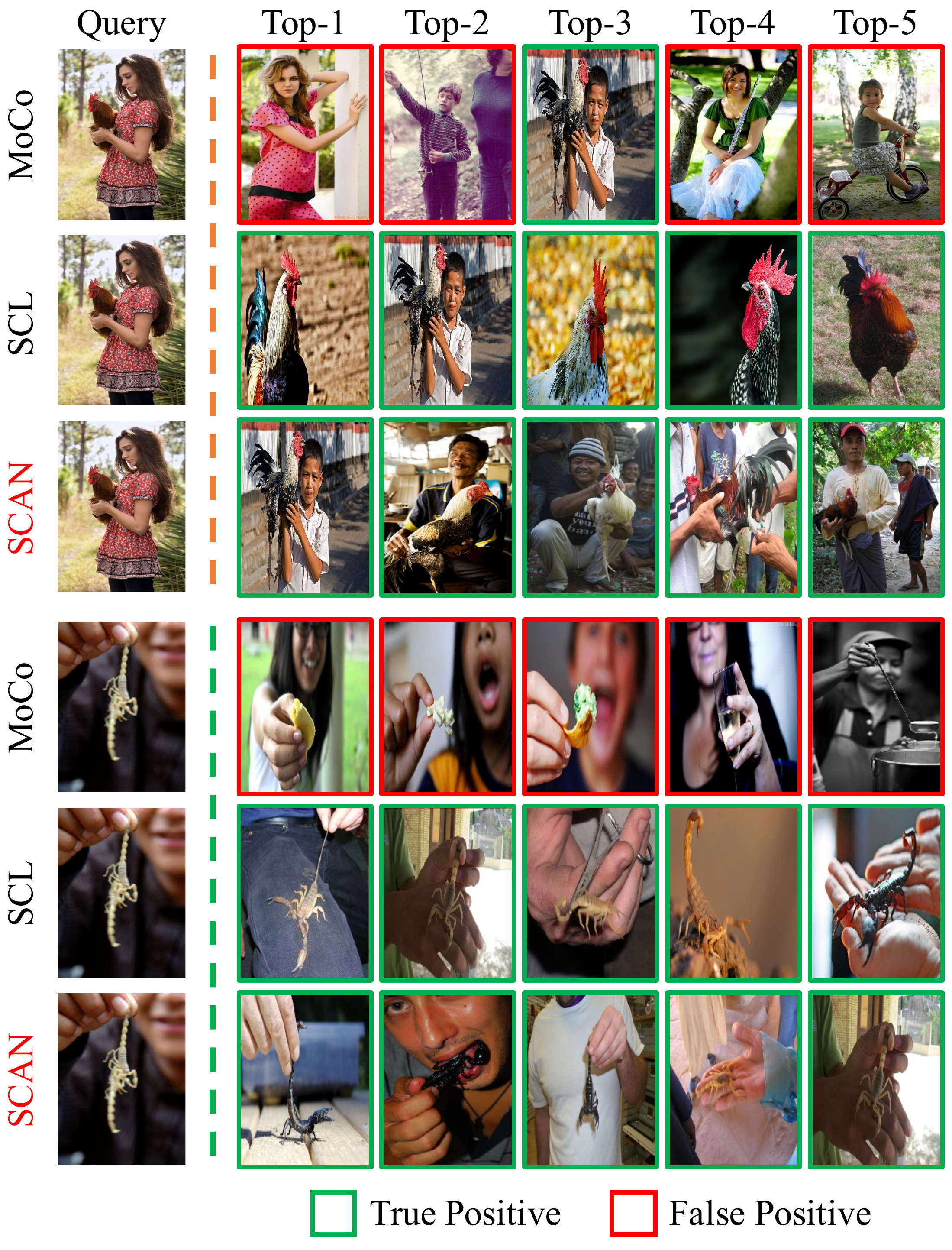}
    \caption{Examples of retrieval results with the features generated by MoCo(v2), SCL, and SCAN. True Positive means that the returned images belong to the same class of the query, and False Positive implies different classes.}
    \label{fig:fig5}
\end{figure}

\subsubsection{Analysis of Representation Learning in SCAN} Though SCAN achieves excellent performance on different downstream tasks, there remains a question of whether it learns more useful knowledge compared with the self-supervised and fully-supervised learning methods. To evaluate this, we conduct \textbf{visualization analysis} on ImageNet-1K. The retrieval results with the features generated by MoCo, SCL, and SCAN are shown in Fig.~\ref{fig:fig5}. Let us take the above case as the example. MoCo mainly focuses on the overall layout (a person at the center of the image) but SCL fits the semantic labels (\textit{i.e.}, the rooster). Both of them are prone to lose part of image information. In comparison, SCAN achieves a balance between them and retrieves the images that are similar to the query in both semantics and appearance. We argue that this property is helpful in deploying the pre-trained model to the downstream tasks.

Additionally, to \textbf{quantitatively analyze} how SCAN enhances the semantic discrimination ability by pulling the positive neighbors, we conduct comparisons with recent self-supervised methods in linear evaluation task on ImageNet-1K. As shown in Tab.~\ref{tab:7}, SCAN achieves $75.0\%$ on Top-1 accuracy, with $3.9\%$ improvement compared with MoCo-v2. Moreover, compared with other state-of-the-art self-supervised methods, SCAN still shows competitive performance.

For that SCAN leverages additional labels compared with self-supervised learning methods, we acknowledge that the above comparison experiment on ImageNet-1K is somewhat unfair. To better evaluate that SCAN indeed learns more transferrable semantic knowledge, we further do analysis on downstream tasks. For the detection task in COCO, researchers have divided the annotated objects into three types according to their object size,~\emph{i.e.}, the small, middle, and large objects, respectively. Therefore, while detecting the large objects, the requirement of the semantic discrimination ability is relatively weaker than that while detecting the small and middle objects. Consequently, as shown in Tab.~\ref{tab:8}, SCAN indeed achieves much better performance in detecting the small and middle objects, but slightly worse performance in detecting the large objects compared with MoCo-v2. These results verify that SCAN improves visual representations by integrating the semantic information into the appearance features, and this mainly owes to the semantic labels properly leveraged.

\begin{table}[]
    \centering
    \caption{ Accuracy of different pre-training methods under linear evaluation on ImageNet-1K with ResNet-50 backbone. * represents the semantic labels are used.}
    \begin{tabular}{l|c|c}
        \hline
        Methods & Top-1 Acc.($\%$) & Top-5 Acc.($\%$) \\
        \hline\hline
        CMC~\cite{tian2019contrastive} & 66.2 & 87.0\\
        MoCo-v1~\cite{he2020momentum} & 60.8 & - \\
        MoCo-v2~\cite{chen2020improved}& 71.1 & -\\
        SimCLR~\cite{chen2020simple} & 69.3 & 89.0\\
        InfoMin~\cite{tian2020makes} & 73.0 & 91.1\\
        BYOL~\cite{grill2020bootstrap} & 74.3 & 91.6\\
        \hline
        SCAN* & \textbf{75.0} & \textbf{91.6}\\
        \hline
    \end{tabular}
    \label{tab:7}
\end{table}

\begin{table}[]
    \centering
    \caption{Analysis of utilizing different pre-training models to fine-tune Mask R-CNN with the R50-C4 backbone and 2$\times$ schedule on the COCO detection task. AP$^\mathrm{bb}_\mathrm{S}$, AP$^\mathrm{bb}_\mathrm{M}$, and AP$^{bb}_\mathrm{L}$ represent the AP metric on detecting the small, middle, and large objects according to the COCO standard.}

    \begin{tabular}{l|c|ccc}
        \hline
         Methods & Data & AP$^\mathrm{bb}_\mathrm{S}$ & AP$^\mathrm{bb}_\mathrm{M}$ &AP$^\mathrm{bb}_\mathrm{L}$\\
        \hline\hline
        MoCo-v2 & IN & 23.8 & 45.6 & \textbf{56.5}\\
        \hline
        SCAN  & IN & \textbf{24.3} & \textbf{46.7} & 56.3\\
        \hline
    \end{tabular}
    \label{tab:8}
\end{table}

\section{Conclusions}
In this paper, we present Supervised Contrastive Adjustment in Neighborhood (SCAN), the first work showing that making good use of semantic labels in contrastive learning can benefit downstream tasks. By simply pulling each query image and its \textit{positive neighbors} together while pushing other images away, SCAN can maximally learn knowledge from both appearance and semantics. Extensive experiments on various downstream tasks clearly demonstrated the effectiveness of SCAN. Importantly,
our research reveals the potential of making \textit{proper} use of supervision labels in assisting the self-supervised visual representation learning and we believe this direction is important for the community.

{\small
\bibliographystyle{ieee_fullname}
\bibliography{egbib}
}

\end{document}